\documentclass[10pt,twocolumn,letterpaper]{article}

\usepackage{iccv}              

\usepackage{multirow}
\usepackage{amssymb}
\usepackage{pifont}
\usepackage{xcolor}

\newcommand{\myparagraph}[1]{\vspace{-6pt}\paragraph{#1}}

\def\tokenizerfull{Deep Compression Hybrid Tokenizer\xspace}
\def\tokenizershort{DC-HT\xspace}

\definecolor{mydarkred}{rgb}{0.8,0.02,0.02}

\definecolor{lightblue}{rgb}{0.9, 0.95, 1}

\def\modelterm{DC-AR\xspace}

\makeatletter
\def\blfootnote#1{\xdef\@thefnmark{}\@footnotetext{\scriptsize #1}}
\makeatother

\definecolor{iccvblue}{rgb}{0.21,0.49,0.74}
\usepackage[pagebackref,breaklinks,colorlinks,allcolors=iccvblue]{hyperref}

\def\paperID{14643} 
\def\confName{ICCV}
\def\confYear{2025}

\title{\modelterm: Efficient Masked Autoregressive Image Generation with \\ Deep Compression Hybrid Tokenizer}

\author{
  Yecheng Wu \quad
  Junyu Chen \quad
  Zhuoyang Zhang \quad
  Enze Xie \quad
  Jincheng Yu \quad \\
  Junsong Chen \quad
  Jinyi Hu \quad
  Yao Lu \quad
  Song Han \quad
  Han Cai \\
  NVIDIA \\
  \url{https://github.com/dc-ai-projects/DC-AR}
}

\begin{document}

\twocolumn[{
\maketitle
\begin{center}
    \captionsetup{type=figure}
    \vspace{-10pt}
    \includegraphics[width=\textwidth]{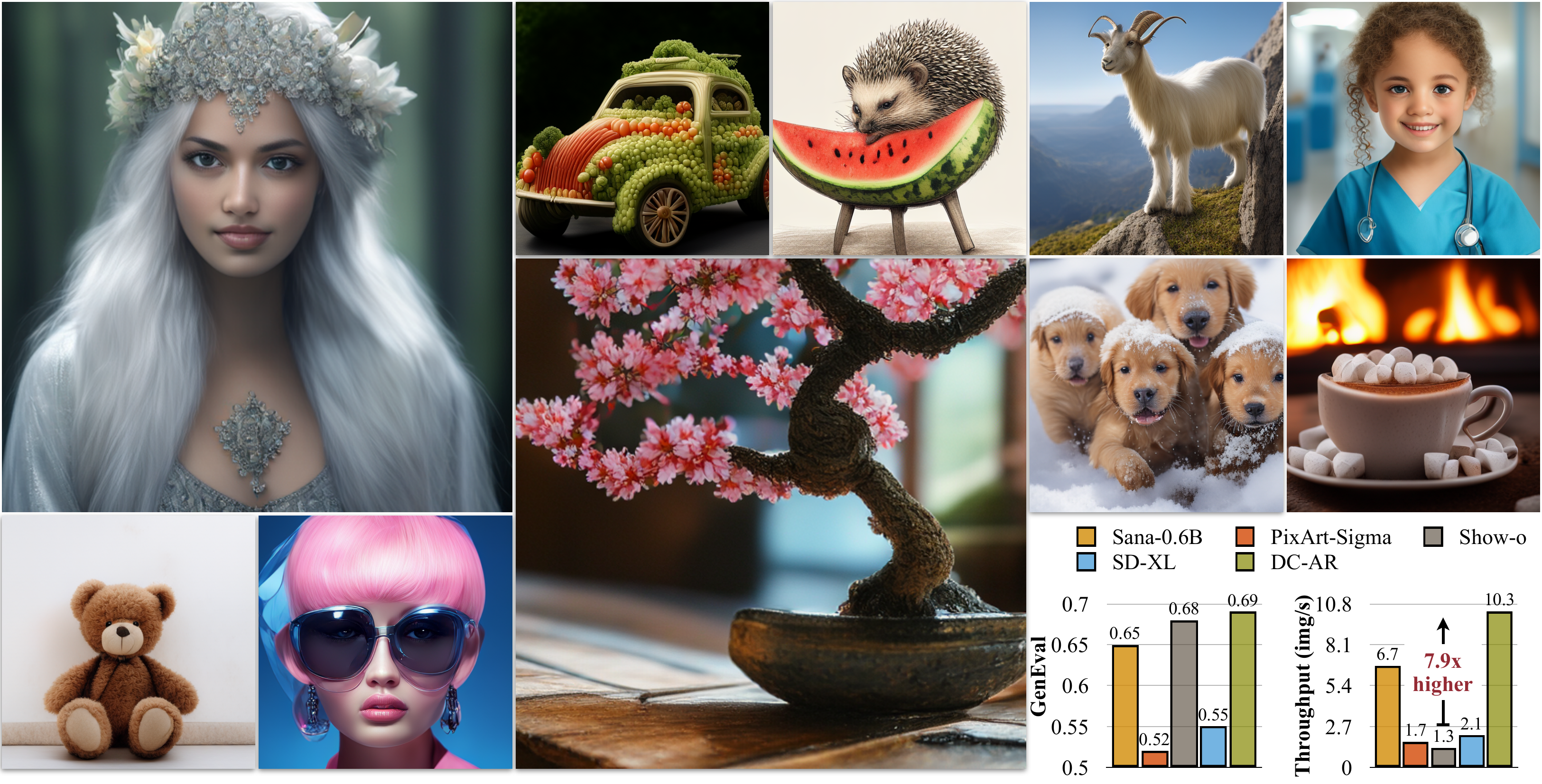}
    \captionof{figure}{\textbf{\modelterm} is a cutting-edge masked autoregressive text-to-image generation framework. In comparison to other leading masked autoregressive models and diffusion models, \modelterm delivers \textbf{1.5-7.9}$\times$ higher throughput and \textbf{2.0-3.5}$\times$ lower latency, all while achieving state-of-the-art quality on text-to-image generation benchmarks.}
\end{center}
}]

\blfootnote{Correspondence to: Han Cai (\texttt{hcai@nvidia.com}).}

\begin{abstract}
We introduce \modelterm, a novel masked autoregressive (AR) text-to-image generation framework that delivers superior image generation quality with exceptional computational efficiency. Due to the tokenizers' limitations, prior masked AR models have lagged behind diffusion models in terms of quality or efficiency. 
We overcome this limitation by introducing \tokenizershort— a deep compression hybrid tokenizer for AR models that achieves a 32$\times$ spatial compression ratio while maintaining high reconstruction fidelity and cross-resolution generalization ability. 
Building upon \tokenizershort, we extend MaskGIT and create a new hybrid masked autoregressive image generation framework that first produces the structural elements through discrete tokens and then applies refinements via residual tokens.
\modelterm achieves state-of-the-art results with a gFID of \textbf{5.49} on MJHQ-30K and an overall score of \textbf{0.69} on GenEval, while offering \textbf{1.5-7.9}$\times$ higher throughput and \textbf{2.0-3.5}$\times$ lower latency compared to prior leading diffusion and autoregressive models. 
\end{abstract}    
\section{Introduction}
\label{sec:intro}

The remarkable success of large language models (LLMs) has significantly propelled advancements in artificial intelligence. Central to this success are autoregressive models, which have gained widespread recognition due to their exceptional generalizability and robust scaling properties. Although primarily employed for natural language processing (NLP) tasks, autoregressive models have also been effectively adapted to image generation, leading to notable progress in the field \cite{fan2024fluid,tang2024hart,han2024infinity,wang2024emu3}. 

In contrast to diffusion models \cite{esser2024scaling,xie2024sana,flux_github,liu2024playground,pernias2024wrstchen} which have dominated the current era, autoregressive models have recently garnered significant attention in image synthesis due to their unique advantages, such as enhanced connections with vision-language models \cite{team2023gemini,openai2024gpt4o,xie2024show,wu2024vila,wu2024janus,jang2024lantern}. Typically, these approaches employ a visual tokenizer to transform images from pixel space into discrete tokens via vector quantization. Then, the models can process these visual tokens analogously to text tokens.

Previous research has explored various paradigms for autoregressive image generation, with masked autoregressive models emerging as an auspicious approach. Inspired by BERT \cite{devlin2019bert} from the field of NLP, these models generate image token sequences through a progressive unmasking process. Unlike the popular GPT-like sequential next-token prediction paradigm, masked autoregressive models allow multiple tokens to be generated simultaneously at each step, significantly reducing the number of sampling steps and improving generation efficiency. Additionally, the masking and unmasking mechanism inherent to these models makes them naturally well-suited for image editing tasks. Since their introduction in MaskGIT \cite{chang2022maskgit}, extensive research has been conducted to extend this method, leading to notable advancements and successes \cite{yu2023magvit,yu2022scaling, kondratyuk2024videopoet,chang2023muse,villegas2022phenaki,xie2024show,chen2024maskmamba,bai2024meissonic,webermaskbit,li2023mage}.

However, compared to diffusion models, which utilize continuous tokens to represent images, autoregressive models face significant challenges for achieving competent efficiency and quality from the image tokenization process. The current standard practice for image tokenization employs an autoencoder with spatial reduction ratios of 8$\times$ or 16$\times$, where an image of size $256\times 256$ is converted into $32\times 32$ or $16\times 16$ token, respectively. However, further reducing token numbers for improved efficiency remains critical, especially considering that computational costs increase when generating high-resolution images. 

While DC-AE \cite{chen2024deep} has successfully achieved a high spatial compression ratio (e.g., 32$\times$) for continuous tokenizers, it remains challenging to build a high-compression tokenizer that we can use for masked autoregressive models. In our experiments, we find directly applying DC-AE to the discrete tokenizer leads to awful reconstruction quality (Table~\ref{tab:ablation_hybrid}). In parallel, some recent works \cite{yu2025image,kim2025democratizing} adopt an alternative method to reduce tokens for discrete tokenizers by employing transformers to convert images into compact 1D latent sequences. Despite their impressive results, such designs break the spatial correspondence between 2D image patches and tokens, making them unable to generalize across different resolutions. They cannot reuse weights of models trained on lower resolutions, making their training cost-prohibitive for high-resolution image synthesis. 

This work introduces \modelterm, an efficient masked autoregressive framework for text-to-image generation. \modelterm incorporates \tokenizershort, a single-scale 2D tokenizer that achieves a spatial reduction ratio of 32$\times$ while maintaining competitive reconstruction quality through hybrid tokenization. Hybrid tokenization, initially proposed by HART \cite{tang2024hart}, is a technique designed to bridge the performance gap between discrete and continuous tokenizers. However, unlike HART, which employs a multi-scale tokenizer with a reduction ratio of 16$\times$, training a single-scale tokenizer with a 32$\times$ reduction ratio using conventional methods often results in suboptimal reconstruction quality. 

\begin{figure*}[t]
    \centering
     \includegraphics[width=0.9\textwidth]{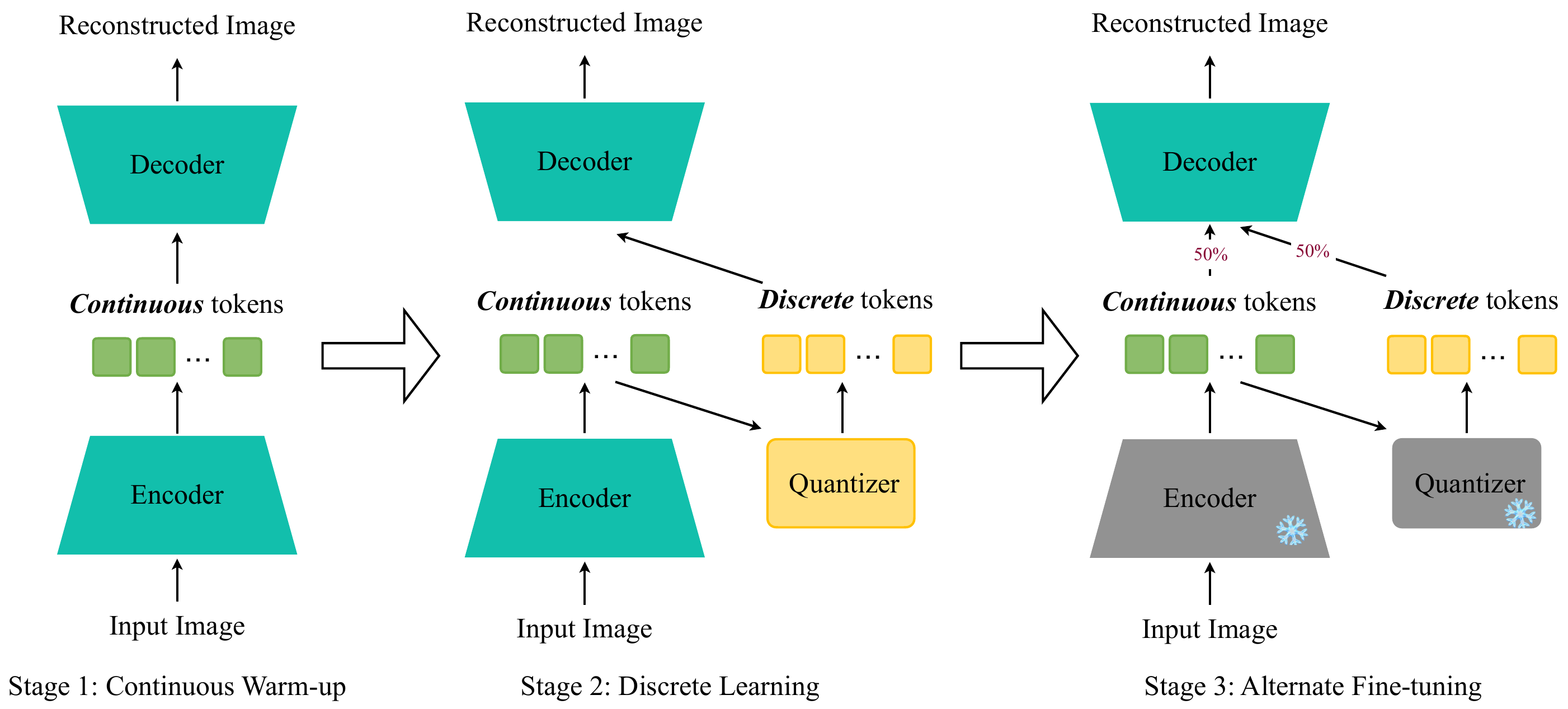}
    \caption{\textbf{Illustration of Our Three-Stage Adaptation Training Strategy for \tokenizershort.} } 
    \label{fig:tokenizer}
\end{figure*}

To overcome these limitations, we propose a three-stage adaptation strategy (Figure~\ref{fig:tokenizer}) to train our tokenizer effectively. This strategy enables \tokenizershort to achieve performance comparable to 1D compact tokenizers at the same compression level while capable of generalizing across resolutions. 
Building on this tokenizer, we develop our text-to-image framework, which utilizes a hybrid coarse-to-fine generation process based on MaskGIT. We train our models with cross-entropy loss for discrete tokens and diffusion loss for residual tokens. At inference time, our framework first generates all discrete tokens through the unmasking process with a transformer model and then produces residual tokens via the denoising schedule of a lightweight diffusion head. These tokens are combined and de-tokenized to create the final output images, guided by the provided textual inputs.

\begin{figure*}[t]
    \centering
\includegraphics[width=0.9\textwidth]{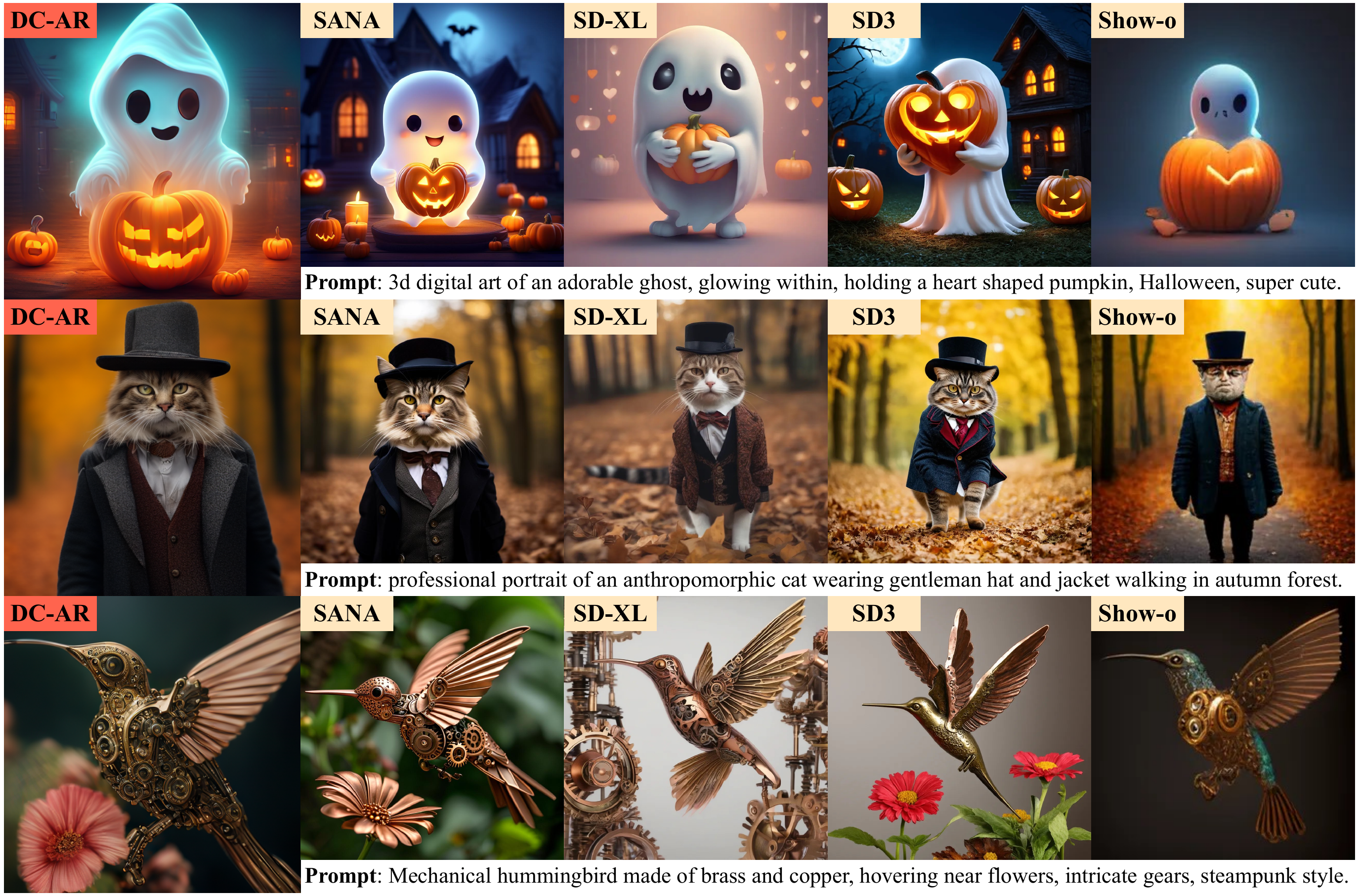}
    \caption{\textbf{Qualitative Comparison of Text-to-Image Generation Results Between \modelterm and Other Generative Models.}}
    \label{fig:generation}
\end{figure*}

We demonstrate the effectiveness of \modelterm across a diverse set of benchmarks. Our tokenizer achieves an rFID of \textbf{1.60} on ImageNet 256$\times$256 \cite{deng2009imagenet}, a performance comparable to 1D tokenizers with the same compression ratio. For text-to-image tasks, \modelterm achieves state-of-the-art results, with a gFID of \textbf{5.49} on MJHQ-30K and an overall score of \textbf{0.69} on GenEval. Additionally, \modelterm delivers \textbf{1.5-7.9}$\times$ higher throughput and \textbf{2.0-3.5}$\times$ lower latency compared to leading diffusion and masked autoregressive models, showcasing its great advantage as an efficient text-to-image generation framework. We summarize our contributions as follows:
\begin{itemize}[leftmargin=*]
\item We build \tokenizershort that significantly reduces the token number to boost AR models' efficiency while maintaining competitive reconstruction quality and cross-resolution generalization capacity. 
\item We introduce an effective three-stage adaptation strategy to improve \tokenizershort's reconstruction quality. 
\item We build \modelterm based on \tokenizershort. \modelterm delivers a significant efficiency boost over prior masked AR models and diffusion models while providing better image generation quality. 
\end{itemize}

\section{Related Work}
\label{sec:related}

\myparagraph{Image Tokenizer.} Since directly learning representation and generation in pixel space is computationally expensive and challenging, modern methods employ image tokenization to convert images into a latent space. These approaches primarily fall into two categories: strategies based on continuous latent token, pioneered by latent diffusion models \cite{rombach2022high} for diffusion models \cite{podellsdxl,peebles2023scalable,bao2023all,cai2024condition}, and methods based on discrete token~\cite{van2017neural,esser2021taming} for autoregressive models \cite{esser2021taming,chang2022maskgit}. 

Traditional tokenizers use 8$\times$ or 16$\times$ spatial compression ratios. Recent research suggests that images can be effectively represented with fewer tokens, leading to significant efficiency gains in the generation process \cite{chen2024deep,xie2024sana,yu2025image}. DC-AE \cite{chen2024deep}, featuring 32$\times$ and higher compression ratios for continuous latents, follows the conventional 2D spatial tokenizer design. TiTok \cite{yu2025image} established a new tokenization paradigm by employing transformer-based models to generate compact 1D global image representations with as few as 32 tokens, inspiring numerous subsequent works \cite{kim2025democratizing,zha2024language,bachmann2025flextok,chen2024softvq,chen2025masked}. However, while practical at fixed resolutions, such compact 1D representations struggle to generalize across different resolutions—an essential advantage of traditional 2D spatial tokenizers. In this work, we adopt the established 2D spatial tokenization approach and develop a hybrid tokenizer with a 32$\times$ compression ratio for efficient autoregressive visual generation.

\myparagraph{Autoregressive Image Generation.} While diffusion models currently dominate text-to-image generation \cite{xie2024sana,xie2025sana,esser2024scaling,team2024kolors,chen2024pixart-alpha,chen2024pixart-delta,chen2024pixart-sigma,li2024playground,team4introducing,maexploring,liu2024playground}, autoregressive approaches have also demonstrated significant potentials \cite{fan2024fluid,han2024infinity,wang2024emu3,chen2025janus}. Autoregressive visual generation primarily follows three paradigms. The first, exemplified by VQGAN \cite{crowson2022vqgan} and its successors \cite{ramesh2021zero,ding2021cogview,ding2022cogview2,liu2024world,sun2024autoregressive,crowson2022vqgan,gafni2022make,wang2024emu3,liu2024lumina,wu2024vila,jang2024lantern,wang2024parallelized,li2023mage}, adopts a GPT-like sequential next-token prediction approach. The second, pioneered by MaskGIT \cite{chang2022maskgit} and extended by various works \cite{chang2023muse,villegas2022phenaki,xie2024show,chen2024maskmamba,bai2024meissonic,li2025autoregressive,fan2024fluid,kim2025democratizing,yu2022scaling}, employs a BERT-like masked autoregressive process. The third, introduced by VAR \cite{tian2025visual} and further developed in subsequent research \cite{ma2024star,zhang2024var,li2024controlvar,yao2024car,han2024infinity,tang2024hart,qu2024tokenflow,chen2024collaborative}, generates images through a progressive next-scale refinement process. Our work builds upon the MaskGIT paradigm, extending it into a hybrid generation framework for efficient text-to-image synthesis.

\section{Method}
\label{sec:method}

In this section, we first introduce \tokenizerfull (\tokenizershort), a 2D tokenizer for autoregressive generation that achieves a spatial compression ratio of 32$\times$ and our three-stage adaptation training strategy to guarantee its decent reconstruction performance. Next, we present \modelterm, an efficient masked autoregressive text-to-image generation framework built upon \tokenizershort.

\begin{figure*}[t]
    \centering
\includegraphics[width=0.9\textwidth]{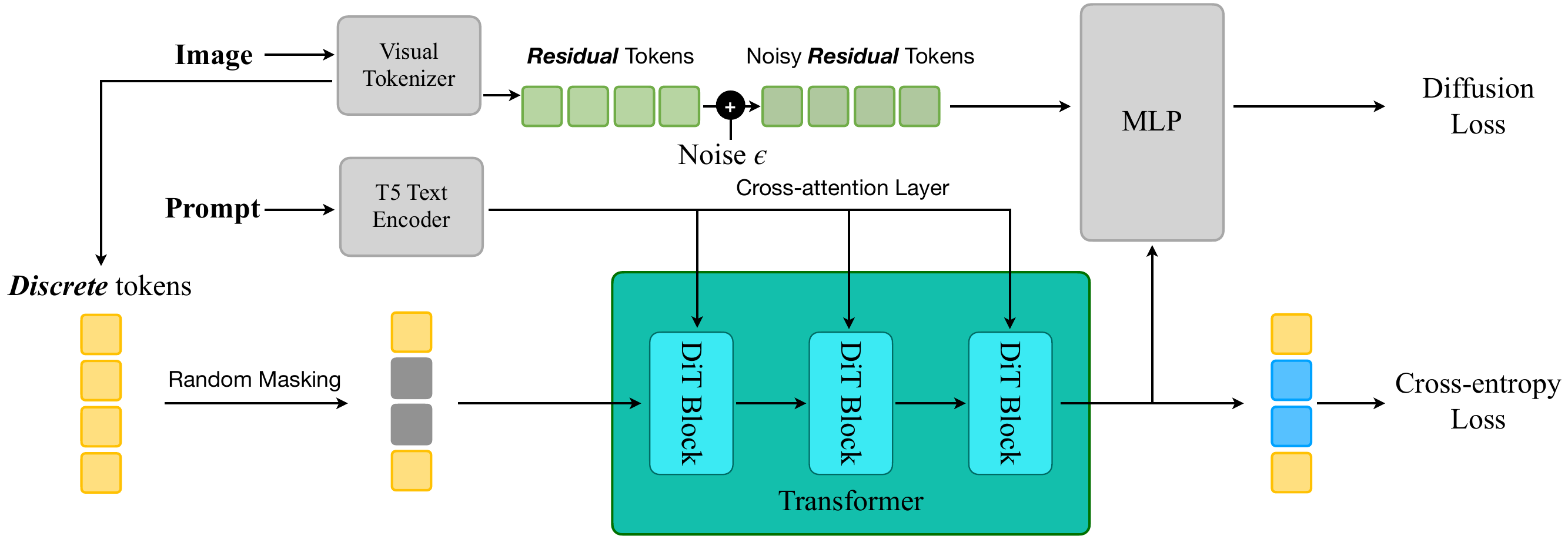}
    \caption{\textbf{Training Workflow of \modelterm.} Our visual tokenizer first decomposes the input image into discrete and residual continuous tokens. As a hybrid generation framework, we design \modelterm to model both types of tokens effectively. We use a cross-entropy loss through the mask-prediction objective to learn discrete tokens. Simultaneously, the hidden states generated by the transformer serve as conditions for an MLP, which predicts the residual tokens using a diffusion loss. The input prompt is injected into the transformer via cross-attention layers to incorporate textual guidance.}
    \label{fig:generator_train}
\end{figure*}

\begin{figure*}[t]
    \centering
\includegraphics[width=0.95\textwidth]{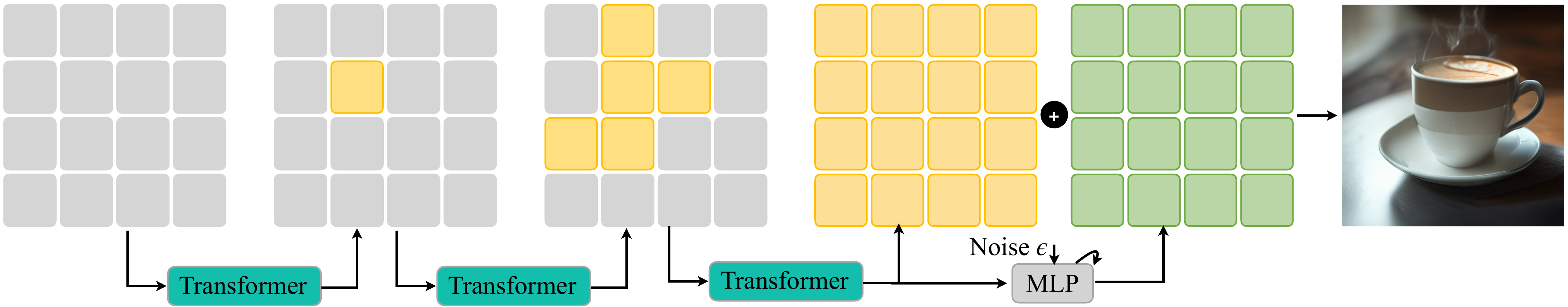}
    \caption{\textbf{Inference Workflow of \modelterm.} The process begins with a fully masked state, where we progressively predict discrete tokens using an unmasking schedule. Once we generate all discrete tokens, the final hidden states from the transformer are utilized as conditions for the diffusion head, which predicts the residual tokens through a denoising schedule. The discrete and residual tokens are combined through summation and decoded to produce the final output image. This two-stage approach ensures structural coherence and fine-grained detail in the generated image.} 
    \label{fig:generator_inference}
\end{figure*}

\subsection{\tokenizerfull}
\label{subsec:tokenizer}

While existing 1D tokenizers used in autoregressive modeling can achieve high compression ratios, they discard the 2D spatial correspondence between pixel patches, limiting their generalizability across different resolutions. To address this, we adopt the 2D discrete tokenizer framework, which includes a CNN-based encoder \(\textit{Enc}\), a CNN-based decoder \(\textit{Dec}\), and a vector-quantized (VQ) quantizer \(\textit{Quant}\). We adopt the same model architecture \cite{cai2023efficientvit} as DC-AE \cite{chen2024deep}, as it delivers state-of-the-art reconstruction quality in continuous tokenization with high compression ratios. Additionally, we find that discrete tokenization is highly sensitive during the codebook training process. Under a high spatial compression ratio, we observe directly training the 2D discrete deep compression tokenizer results in poor reconstruction quality (Table~\ref{tab:ablation_hybrid}). In this work, we propose to enhance it with hybrid tokenization and a three-stage adaptation strategy to mitigate quality loss. 

\myparagraph{Hybrid Tokenization.} Given an input image \(\mathbf{I}\), the reconstruction process can proceed via either a \textit{discrete path} or a \textit{continuous path}. In the \textit{discrete path}, the input image \(\mathbf{I}\) is first compressed by the CNN encoder \(\textit{Enc}\) into a latent representation of continuous tokens \(\mathbf{Z} = \textit{Enc}(\mathbf{I})\), followed by the VQ process \(\mathbf{Z}_q = \textit{Quant}(\mathbf{Z})\) to obtain discrete tokens. The discrete tokens \(\mathbf{Z}_q\) are then passed through the CNN decoder \(\textit{Dec}\) to reconstruct the image \(\hat{\mathbf{I}} = \textit{Dec}(\mathbf{Z}_q)\). The reconstruction loss and GAN loss between \(\mathbf{I}\) and \(\hat{\mathbf{I}}\) are computed to train the tokenizer. In the \textit{continuous path}, the quantization step is skipped, and the continuous latent \(\mathbf{Z}\) is directly fed into the decoder \(\textit{Dec}\) to obtain the reconstructed image \(\hat{\mathbf{I}} = \textit{Dec}(\mathbf{Z})\). According to HART, a critical property of the hybrid tokenizer for successful generation is its ability to decode both continuous tokens \(\mathbf{Z}\) and discrete tokens \(\mathbf{Z}_q\) effectively. It ensures that the two types of tokens remain sufficiently similar from the decoder’s perspective, facilitating easier modeling of their residual tokens, defined as \(\mathbf{Z}_r = \mathbf{Z} - \mathbf{Z}_q\), during the generation process.

\vspace{-0.5cm}
\myparagraph{Three-stage Adaptation Training Strategy.} Hybrid tokenization alone cannot fully resolve the reconstruction quality drop, as it faces the inherent conflict between discrete and continuous latent spaces. We find directly applying the alternate training strategy from HART \cite{tang2024hart} leads to unsatisfactory reconstruction results. We propose a Three-Stage Adaptation Training Strategy to tackle this challenge. The detailed training pipeline is illustrated in Figure~\ref{fig:tokenizer}.

The first stage, denoted as \emph{continuous warm-up}, focuses solely on the \textit{continuous path}. This stage is relatively short and aims to initialize the encoder with weights suitable for the reconstruction task. The second stage, denoted as \emph{discrete learning}, activates only the \textit{discrete path}. Here, the goal is to train the tokenizer to learn a stable latent space and enable it to reconstruct images effectively. The final stage, \emph{alternate fine-tuning}, randomly selects either the \textit{continuous path} or the \textit{discrete path} for each image with equal probability (50\%) to fine-tune the tokenizer. During this stage, the encoder and quantizer are frozen, and only the decoder is fine-tuned. This stage ensures that the decoder can effectively handle both continuous and discrete tokens.

By breaking down the training process into these three stages, our strategy effectively addresses the abovementioned issues, improving the rFID from 1.92 to 1.60 and discrete-rFID (rFID for the discrete path) from 6.18 to 5.13. 

\subsection{Hybrid Masked Autoregressive Model}
\label{subsec:generator}

To fully harness the capabilities of \tokenizershort, we build \modelterm, a masked autoregressive framework designed to generate high-resolution images under textual guidance efficiently. Figure~\ref{fig:generator_train} illustrates our general framework. A text model extracts textual embeddings from input prompts, which are then integrated into transformer blocks via cross-attention to provide textual guidance. During training, we mask a random subset of discrete tokens and train the transformer model to predict these masked tokens using a cross-entropy loss. Simultaneously, the hidden states produced by the transformer model serve as conditional inputs for predicting the residual tokens through a lightweight diffusion MLP head, optimized using a diffusion loss \cite{li2025autoregressive}. 

\begin{table*}[!t]
    \small\centering\setlength{\tabcolsep}{3pt}
    \begin{tabular}{lccccccc}
    \toprule
    \multirow{2}{*}{Tokenizer} & \multirow{2}{*}{Type} & Resolution & \multirow{2}{*}{\# Tokens} & \multirow{2}{*}{rFID$\downarrow$} & \multirow{2}{*}{PSNR$\uparrow$} & \multirow{2}{*}{SSIM$\uparrow$} & \multirow{2}{*}{LPIPS$\downarrow$} \\
    & & Generalizable? & & & & \\
    \midrule\midrule
    \textbf{ImageNet 256$\times$ 256} & & & & & & & \\
    \midrule
    TiTok \cite{yu2025image} & 1D-Discrete & \ding{55} & 64 & 1.70 & 17.06 & 0.4021 & 0.3840 \\
    TA-TiTok (VQ) \cite{kim2025democratizing} & 1D-Discrete & \ding{55} & 64 & 2.68 & -- & -- & -- \\
    TA-TiTok (KL) \cite{kim2025democratizing}  & 1D-Continuous & \ding{55} & 64 & 1.47 & -- & -- & -- \\
    TexTok$^*$ \cite{zha2024language} & 1D-Continuous & \ding{55} & 64 & 1.53 & 20.10 & 0.5618 & 0.2126 \\
    \midrule
     \textbf{\tokenizershort (Ours)} & 2D-Hybrid & \ding{51} & 64 & 1.60 & 21.50 & 0.5676 & 0.2221 \\
    \midrule
    \midrule
    \textbf{ImageNet 512$\times$ 512} & & & & & & & \\
    \midrule
    TiTok \cite{yu2025image} & 1D-Discrete & \ding{55} & 128 & 1.37 & -- & -- & -- \\
    TexTok$^*$ \cite{zha2024language} & 1D-Continuous & \ding{55} & 128 & 0.97 & 22.27 & 0.6230 & 0.2365 \\
    TexTok$^*$ \cite{zha2024language} & 1D-Continuous & \ding{55} & 256 & 0.73 & 24.45 & 0.6682 & 0.1875 \\
    \midrule
    \textbf{\tokenizershort  (Ours)} & 2D-Hybrid & \ding{51} & 256 & 0.83 & 23.53 & 0.6315 & 0.2236 \\
    \bottomrule
    \end{tabular}
    \caption{\textbf{Image Reconstruction Results on ImageNet.} $^*$ indicates that the results of TexTok were computed on 50K samples from the ImageNet training set instead of the ImageNet validation set. \tokenizershort consistently demonstrates competitive performance compared to other 1D tokenizers across all benchmarks at the same compression ratio while maintaining generalizability across different resolutions. }
    \label{tab:tokenization-main}
\end{table*}

Figure~\ref{fig:generator_inference} demonstrates our inference pipeline. All discrete tokens are predicted iteratively through a progressive unmasking schedule, starting from a fully masked state. Once we generate all discrete tokens, the final hidden states from the transformer are utilized as conditions
for the diffusion head, which predicts the residual tokens through a denoising schedule. We then sum the predicted discrete and residual tokens to obtain the final continuous tokens, which we use to produce the generated image with our decoder. 

A key design choice in our hybrid generation framework is that \textit{only discrete tokens} are involved in the forward process of our transformer model. This approach is grounded in the principle that residual tokens should exclusively serve a refining function without altering the overall structure of the generated images. This design is critical, as empirical evidence \cite{chang2022maskgit} has shown that discrete token-based MaskGIT typically requires as few as eight steps to achieve near-optimal generation performance. In contrast, continuous token-based MAR \cite{li2025autoregressive} needs 64 steps to reach optimal performance (please refer to \cref{subsec:additional_experiments}), leading to significantly higher inference costs. By relying on discrete tokens for the transformer prediction process and residual tokens for refinement, we ensure that our framework maintains the high sampling efficiency of discrete token-based approaches like MaskGIT while achieving superior image generation quality.

As discussed in \cref{subsec:tokenizer}, a key advantage of our 2D spatial tokenizer design over 1D tokenizers is its ability to generalize seamlessly across different resolutions, producing tokens that reside in the same latent space. Leveraging this property, we adopt a two-stage training strategy to train an image generation model for 512$\times$512 images efficiently. First, using a relatively long training schedule, we pre-train our model on 256$\times$256 images. Subsequently, we fine-tune the pre-trained 256$\times$256 model on 512$\times$512 images to obtain the final model, which converges rapidly due to the shared latent space. As demonstrated in \cref{subsec:ablation}, this training pipeline reduces GPU hours by at least 1.9$\times$ compared to training the 512$\times$512 model from scratch, significantly enhancing training efficiency.

\section{Experiments}
\label{sec:experiments}

\subsection{Setups}
\label{subsec:setup}

\myparagraph{Models.} For the tokenizer, we adopt the DC-AE-f32c32 architecture from \citet{chen2024deep}, featuring a spatial compression ratio of 32$\times$ and a latent channel size of 32. We configure the codebook to $N = 16384$. For the generator, we utilize the PixArt-$\alpha$ \cite{chen2024pixart-alpha} architecture for our transformer model, with its adaptive norm layers removed. It consists of 28 layers and a width of 1152, amounting to 634M parameters. The diffusion head comprises 6 MLP layers, totaling 37M parameters. To ensure computational efficiency and accessibility for research environments, we employ T5-base \cite{raffel2020exploring} as our text encoder, which contains 109M parameters.

\myparagraph{Evaluation and Datasets.} For the tokenizer, we use the training split of ImageNet \cite{deng2009imagenet} as our training dataset, resizing each image to 256$\times$256. To assess the reconstruction performance of the tokenizer, we evaluate reconstruction FID \cite{heusel2017gans} (rFID), peak signal-to-noise ratio (PSNR), structural similarity index measure (SSIM), and learned perceptual image patch similarity (LPIPS) \cite{zhang2018unreasonable} on the ImageNet validation set at resolutions of 256$\times$256 and 512$\times$512. For the text-to-image generator, we employ JourneyDB \cite{sun2023journeydb} and an internal MidJourney-style synthetic dataset, where each data point consists of an image-caption pair, where the captions are generated using VILA1.5-13B \cite{lin2024vila}. To evaluate generation performance, we report generation FID (gFID) on MJHQ-30K \cite{li2024playground} to measure aesthetic quality and the GenEval \cite{ghosh2023geneval} score to quantify the alignment between input prompts and generated images.

\myparagraph{Efficiency Profiling.} We profile the latency and throughput on an NVIDIA A100 GPU. We measure the throughput with a batch size of 16 and the latency with a batch size of 1. We use float16 precision for all cases.

\begin{table*}[!t]
    \small\centering\setlength{\tabcolsep}{3pt}
    \begin{tabular}{lcccc|c|cc}
    \toprule
    \multirow{2}{*}{Method} & \multirow{2}{*}{Type} & \multirow{2}{*}{\#Params} & \multirow{2}{*}{Resolution} & \multirow{2}{*}{\#Steps} & \multirow{2}{*}{gFID$\downarrow$} & Latency & Throughput  \\
    & & & & & & (s) & (images/s) \\
    \midrule
    SDXL \cite{podellsdxl} & Diffusion & 2.6B & 1024 $\times$ 1024 & 20 & 6.63 & 1.4$^*$ & 2.1$^*$ \\
    PixArt-$\alpha$ \cite{chen2024pixart-alpha} & Diffusion & 630M & 512 $\times$ 512 & 20 & 6.14 & 1.2 & 1.7 \\
    PixArt-$\Sigma$ \cite{chen2024pixart-sigma} & Diffusion & 630M & 512 $\times$ 512 & 20 & 6.34 & 1.2 & 1.7 \\
    SD3-medium \cite{esser2024scaling} & Diffusion & 2B & 1024 $\times$ 1024 & 28 & 11.92 & -- & -- \\
    Playground v2.5 \cite{li2024playground} & Diffusion & 2B & 1024 $\times$ 1024 & 20 & 6.09 & -- & -- \\
    Sana-0.6B \cite{xie2024sana} & Diffusion & 590M & 512 $\times$ 512 & 20 & 5.67 & 0.8 & 6.7 \\
    \midrule
    Show-o \cite{xie2024show} & Mask. AR & 1.3B & 512 $\times$ 512 & 12 & 14.59 & 1.1 & 1.3 \\
    TA-TiTok (VQ) \cite{kim2025democratizing} & Mask. AR & 568M & 256 $\times$ 256 & 16 & 7.74 & -- & -- \\
    TA-TiTok (KL) \cite{kim2025democratizing} & Mask. AR & 602M & 256 $\times$ 256 & 32 & 7.24 & -- & -- \\
    \midrule
    \textbf{DC-AR (Ours)}  & Mask. AR & 671M & 512 $\times$ 512 & 12 & \textbf{5.49} & \textbf{0.4} & \textbf{10.3} \\
    \bottomrule
    \end{tabular}
    \caption{\looseness=-1 \textbf{Text-to-Image Generation Results on MJHQ-30K.} We evaluate \modelterm against state-of-the-art diffusion and masked autoregressive models. * denotes that we report the latency and throughput for generating 512$\times$512 images. The results demonstrate that \modelterm achieves comparable image generation quality while offering significant efficiency gains: \textbf{2.0-3.5$\times$} lower latency and \textbf{1.5-7.9$\times$} higher throughput compared to existing methods when generating 512$\times$512 images.}    \label{tab:generation-main}
\end{table*}
\begin{table*}[!t]
    \begin{center}
        \scalebox{0.95}{
            \begin{tabular}{lcc|cccccc|c}
               \toprule
               Method & Type & \#Params & S. Obj & T. Obj & Count & Colors & Position & C. Attri. & Overall \\
               \midrule
               SDXL \cite{podellsdxl} & Diffusion & 2.6B & 0.98 & 0.74 & 0.39 & 0.85 & 0.15 & 0.23 & 0.55  \\
               PixArt-$\alpha$ \cite{chen2024pixart-alpha} & Diffusion & 630M & 0.96 & 0.49 & 0.47 & 0.79 & 0.06 & 0.11 & 0.48  \\
               PixArt-$\Sigma$ \cite{chen2024pixart-sigma} & Diffusion & 630M & 0.98 & 0.59 & 0.50 & 0.80 & 0.10 & 0.15 & 0.52  \\
               SD3-medium \cite{esser2024scaling} & Diffusion & 2B & 0.98 & 0.74 & 0.63 & 0.67 & 0.34 & 0.36 & 0.62 \\
               Playground v2.5 \cite{li2024playground} & Diffusion & 2B & 0.98 & 0.77 & 0.52 & 0.84 & 0.11 & 0.17 & 0.56 \\
               Sana-0.6B \cite{xie2024sana} & Diffusion & 590M & 0.99 & 0.76 & 0.64 & 0.88 & 0.18 & 0.39 & 0.64 \\
               \midrule
               Show-o \cite{xie2024show} & Mask. AR & 1.3B & 0.98 & 0.80 & 0.66 & 0.84 & 0.31 & 0.50 & 0.68 \\
               TA-TiTok (VQ) \cite{kim2025democratizing} & Mask. AR & 568M & 0.98 & 0.57 & 0.46 & 0.80 & 0.11 & 0.25 & 0.53 \\
               TA-TiTok (KL) \cite{kim2025democratizing} & Mask. AR & 602M & 0.99 & 0.57 & 0.36 & 0.80 & 0.11 & 0.29 & 0.52 \\
               Fluid \cite{fan2024fluid} & Mask. AR & 665M & 0.96 & 0.73 & 0.51 & 0.77 & 0.42 & 0.51 & 0.65 \\
               Meissonic \cite{bai2024meissonic} & Mask. AR & 1.0B & 0.99 & 0.66 & 0.42 & 0.86 & 0.10 & 0.22 & 0.54 \\
               \midrule
               \textbf{DC-AR (Ours)}  & Mask. AR & 671M & 1.00 & 0.75 & 0.52 & 0.90 & 0.45 & 0.51 & \textbf{0.69} \\
               \bottomrule
            \end{tabular}
        }
    \end{center}
    \vspace{-10pt}
    \caption{\textbf{Text-to-Image Generation Results on GenEval.} The results demonstrate that \modelterm achieves state-of-the-art performance compared to other methods of similar scale.}
    \label{tab:generation-geneval}
\end{table*}

\subsection{Main Results}
\label{subsec:results}

\myparagraph{Image Tokenization.} The quantitative results presented in \cref{tab:tokenization-main} demonstrate that \tokenizershort achieves reconstruction performance on par with 1D compact tokenizers while maintaining high compression ratios. Notably, \tokenizershort is trained exclusively on 256$\times$256 images yet delivers competitive performance at 512$\times$512 resolution, whereas models for 1D tokenizers require separate training on both 256$\times$256 and 512$\times$512 images. This advantage stems from \tokenizershort's retention of the resolution generalizability inherent to 2D tokenizers, a capability that 1D tokenizers lack.

\myparagraph{Text-to-Image Generation.} We present quantitative text-to-image generation results in \cref{tab:generation-main} and \cref{tab:generation-geneval}. On the MJHQ-30K benchmark, \modelterm achieves the state-of-the-art gFID score of 5.49 compared to leading diffusion models and other masked autoregressive models. Notably, \modelterm accomplishes this with significantly lower inference costs, requiring only 12 steps. For 512\(\times\)512 image generation, \modelterm demonstrates 2.0\(\times\) and 3.5\(\times\) lower latency than Sana-0.6B and SD-XL, respectively, alongside 1.5\(\times\) and 7.9\(\times\) higher throughput than Sana-0.6B and Show-o, respectively. On the GenEval benchmark, \modelterm achieves an overall score of 0.69, matching the state-of-the-art masked autoregressive model Show-o (within 0.01) while being 2\(\times\) smaller in scale. Furthermore, \modelterm outperforms other models of similar scale by at least 0.04. Additionally, we provide qualitative samples comparing the generation results of \modelterm with other advanced models in \cref{fig:generation}. The quantitative and qualitative results underscore \modelterm's capabilities as a cutting-edge text-to-image generation framework with superior efficiency.

\subsection{Ablation Studies and Analysis}
\label{subsec:ablation}

We evaluate the key design choices in \modelterm by examining the following aspects: the effectiveness of our hybrid design over discrete-only baseline, the advantage of our three-stage adaptation strategy for tokenizer training, the training efficiency gains enabled by our resolution generalizable tokenizer for the generator, and the efficiency advantages of our hybrid generation framework in terms of sampling steps.

\myparagraph{Effectiveness of the hybrid design.} Compared to traditional autoregressive methods that rely solely on discrete tokens, the hybrid tokenization and generation design enhances the representation capability of \modelterm, resulting in improved performance. To validate this, we compare \modelterm with a discrete-only baseline that excludes the continuous path, residual tokens, and the diffusion head from \modelterm. We present the results in \cref{tab:ablation_hybrid}. \modelterm outperforms the discrete-only baseline across various comprehensive metrics while incurring only 10\% additional overhead, demonstrating our hybrid formulation's effectiveness. Additionally, qualitative examples demonstrating how the hybrid design enhances generation quality by capturing finer details are provided in \cref{fig:comparsion_discrete} of \cref{subsec:additional_experiments}.

\vspace{-6pt}
\begin{table}[!t]
\begin{center}
    \scalebox{0.9}{
    \begin{tabular}{l|c|c|c|c}
    \hline
     & \multirow{2}{*}{rFID $\downarrow$} & \multirow{2}{*}{gFID $\downarrow$} & \multirow{2}{*}{GenEval $\uparrow$} & Througput  \\
     & & & & (images/s) \\
    \hline
    \modelterm & 1.60 & 5.50 & 0.69 & 10.3 \\
    \hline
    Discrete-only & 5.13 & 6.71 & 0.66 & 11.4 \\
    \hline
    \end{tabular}
}
\end{center}
\vspace{-10pt}
\caption{The hybrid design allows \modelterm to surpass the discrete-only baseline across all evaluation metrics, while incurring only 10\% additional overhead.}
\label{tab:ablation_hybrid}
\vspace{-6pt}
\end{table}

\myparagraph{Three-stage Adaptation Training Strategy.} We evaluate our three-stage adaptation strategy for training the hybrid tokenizer with a compression ratio 32$\times$ against two alternative methods, as shown in \cref{tab:ablation_tokenizer}. The first alternative strategy omits the continuous warm-up stage, which leads to increased difficulty in learning the discrete latent space, resulting in poorer discrete-rFID and continuous-rFID performance for the final tokenizer. The second alternative strategy proceeds directly to the alternate training stage after the continuous warm-up, resembling the alternate fine-tuning stage but with all components trainable. This approach degrades both discrete-rFID and continuous-rFID due to the conflict between the discrete and continuous latent space because the latent space is trainable. Our three-stage strategy effectively addresses these issues, ensuring balanced and optimized performance.

\vspace{-6pt}
\begin{table}[!t]
\begin{center}
    \scalebox{0.9}{
    \begin{tabular}{l|c|c}
    \hline
     & Discrete-rFID $\downarrow$ & rFID $\downarrow$  \\
    \hline
    Discrete Training + &  \multirow{2}{*}{5.93} & \multirow{2}{*}{1.76} \\
    Alternate Fine-tuning & & \\
    \hline
    Continuous Warm-up + &  \multirow{2}{*}{6.18} & \multirow{2}{*}{1.92} \\
    Alternate Training & & \\
    \hline
    Three-Stage Adaptation & 5.13 & 1.60 \\
    \hline
    \end{tabular}
}
\end{center}
\vspace{-10pt}
\caption{The Three-Stage Adaptation Strategy outperforms alternate training strategies for our hybrid tokenizer.}
\label{tab:ablation_tokenizer}
\vspace{-6pt}
\end{table}

\looseness=-1
\myparagraph{Training Efficiency Benefit.} As discussed in \ref{subsec:generator}, our resolution-generalizable tokenizer enables an effective ``pre-training then fine-tuning" strategy for the 512$\times$512 model. This approach begins with pre-training at 256$\times$256 resolution, followed by fine-tuning at the target 512$\times$512 resolution. In contrast, models with single-resolution tokenizers can only be trained at the target resolution from scratch. 
\cref{tab:ablation_training} quantitatively shows that our strategy reduces training costs by more than 1.9$\times$ when compared to training from scratch while maintaining superior generation quality, as measured by gFID scores.

\begin{table}[!t]
\vspace{-6pt}
\begin{center}
    \scalebox{0.9}{
    \begin{tabular}{l|c|c|c}
    \hline
     & Training Steps & GPU Hours & gFID $\downarrow$  \\
    \hline
    Pre-training + & 200K (256) +  & \multirow{2}{*}{760} & \multirow{2}{*}{5.50} \\
    Finetuning & 50K (512) & & \\
    \hline
    Training & \multirow{2}{*}{200K (512)} & \multirow{2}{*}{1440} & \multirow{2}{*}{6.64} \\
    from scratch &  & & \\
    \hline
    \end{tabular}
}
\end{center}
\vspace{-10pt}
\caption{Our ``pre-training then fine-tuning" strategy reduces the training cost by more than \textbf{1.9$\times$} for training a 512$\times$512 model.}
\label{tab:ablation_training}
\vspace{-6pt}
\end{table}

\begin{figure}[t]
    \centering
     \includegraphics[width=0.4\textwidth]{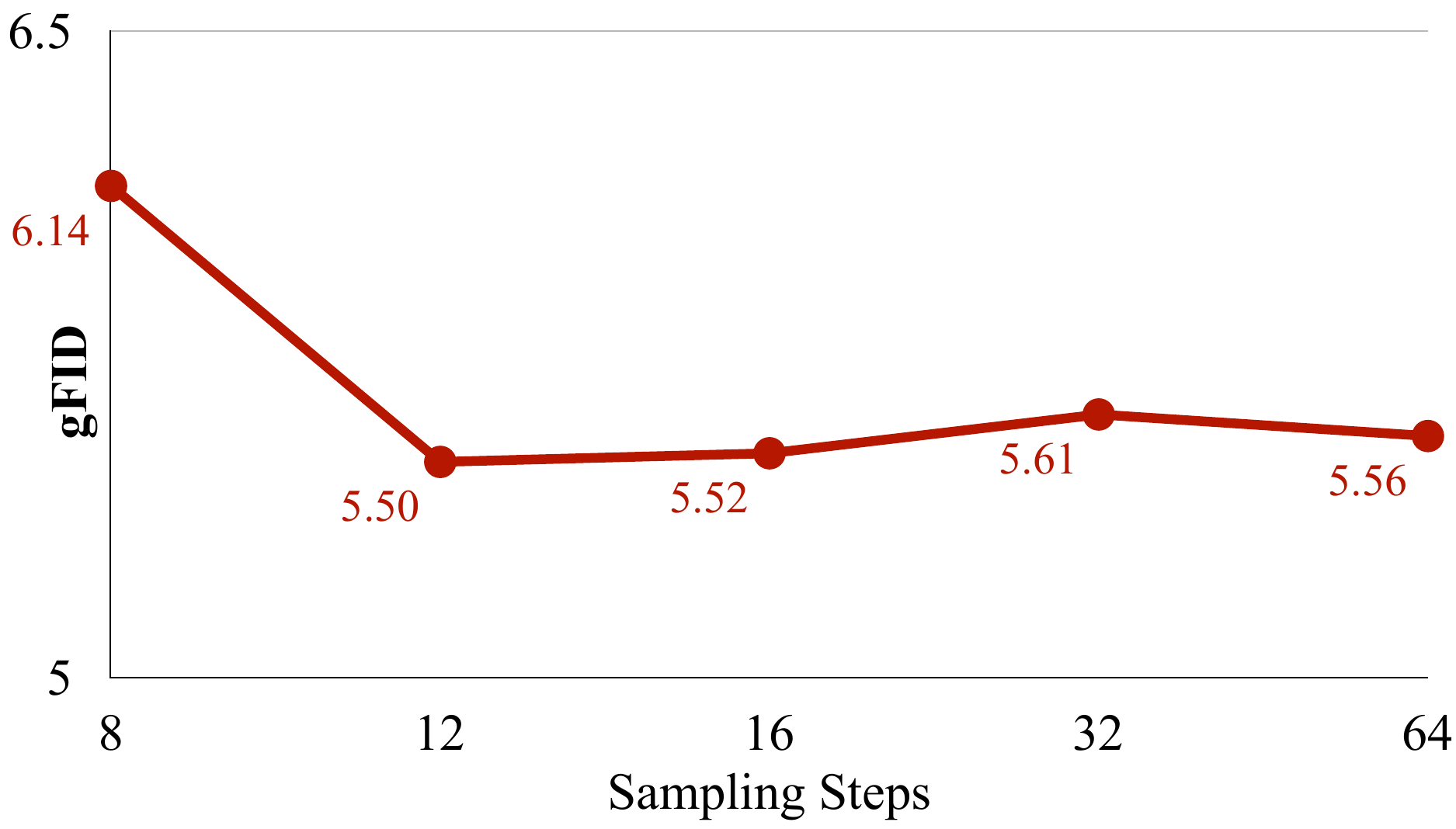}
     \vspace{-5pt}
    \caption{\textbf{Results of gFID under Different Sampling Steps.} Our hybrid generation pipeline, dominated by discrete tokens, allows \modelterm to achieve exceptional quality with just 12 sampling steps. In contrast to MAR-based methods, which require numerous steps to attain optimal performance, this reduced sampling demand grants \modelterm significant efficiency advantages.} 
    \label{fig:ablation_mar}
    \vspace{-5pt}
\end{figure}

\myparagraph{Inference Efficiency Benefit.} \cref{fig:ablation_mar} demonstrates the gFID results of \modelterm under different sampling steps. Our discrete token dominated generation pipeline enables \modelterm to achieve optimal image quality with 12 sampling steps. In contrast, MAR-based \cite{li2025autoregressive} models require numerous steps to reach optimal performance. This reduced demand in sampling steps brings \modelterm great efficiency advantages without compromising generation quality.

\section{Conclusion}
\label{sec:conclusion}

This paper presents \modelterm, a novel and efficient masked autoregressive text-to-image generation framework. While modern diffusion models can leverage tokenizers with high compression ratios, autoregressive models encounter significant challenges when adopting the same approach. To address this, we introduce \tokenizershort, a 2D hybrid tokenizer that achieves a 32$\times$ spatial compression ratio while maintaining exceptional reconstruction fidelity. Building on \tokenizershort, \modelterm is a masked autoregressive generation framework capable of effectively generating discrete and residual tokens to synthesize images. By first predicting structural elements through discrete tokens and then refining details using residual tokens, \modelterm generates high-quality images in just 12 steps. This approach delivers \textbf{1.5-7.9}$\times$ higher throughput and \textbf{2.0-3.5}$\times$ lower latency compared to state-of-the-art diffusion and masked autoregressive models.

\newpage
{
    \small
    \bibliographystyle{ieeenat_fullname}
    \bibliography{iccv}
}
\newpage

\appendix
\newpage
\onecolumn
{
\centering
\Large
\textbf{\thetitle}\\
\vspace{0.5em}Supplementary Material \\
\vspace{1.0em}
}

\setcounter{page}{1}
\setcounter{section}{0}
\renewcommand*{\thesection}{\Alph{section}}

\section{Appendix}

We provide additional information and results in the appendix, as outlined below:

\begin{itemize}
    \item \cref{subsec:ethics}: Ethics Statement, discussing about how we prevent the misuse of \modelterm.
    \item \cref{subsec:implementation}: Implementation Details, including the training hyper-parameters for tokenizer and generator, inference hyper-parameters for generator.
    \item \cref{subsec:text-to-image}: Additional text-to-image generation of \modelterm and other popular methods. 
    \item \cref{subsec:qualitative_comparison}: Qualitative comparison between \modelterm and the discrete-only baseline.
    \item \cref{subsec:additional_experiments}: Additional experiments to help clarfiy the advantages of \modelterm.
\end{itemize}

\subsection{Ethics Statement}
\label{subsec:ethics}

The misuse of generative AI for creating NSFW (not safe for work) content continues to be a critical concern within the community. To address this, we have integrated \modelterm with ShieldGemma-2B \cite{zeng2024shieldgemma}, a robust LLM-based safety check model. In our implementation, user prompts are first evaluated by the safety check model to detect NSFW content, including harmful, abusive, hateful, sexually explicit, or otherwise inappropriate material targeting individuals or protected groups. If a prompt is deemed safe, it is forwarded to \modelterm for image generation. If not, the prompt is rejected and replaced with a default prompt (``A red heart"). Through rigorous testing, we have demonstrated that our safety check model effectively filters out NSFW prompts under strict thresholds, ensuring that our pipeline does not produce harmful AI-generated content.

\subsection{Implementation Details}
\label{subsec:implementation}

\looseness=-1
\cref{tab:training_tokenizer} and \cref{tab:training_generator} present the hyper-parameters used for training the tokenizer and generator, respectively. For image generation, we employ the following sampling hyper-parameters: a randomized temperature of 4.5, a CFG (Classifier-Free Guidance) scale of 4.5, a constant CFG schedule, 12 sampling steps for discrete tokens, and 20 diffusion steps for residual tokens.

\begin{table}[htbp]
\begin{center}
\begin{minipage}{0.42\textwidth} 
    \centering
    \scalebox{1.0}{
    \begingroup
    \fontsize{12}{14}\selectfont
    \begin{tabular}{l|c}
    \hline
    Hyper-parameters & Configuration \\
    \hline
    optimizer & Adam \\
    $\beta_1$ & 0.9 \\
    $\beta_2$ & 0.95 \\
    discriminator loss weight & 0.5 \\
    perceptual loss weight & 1.0 \\
    $L_1$ loss weight & 0.0 \\
    $L_2$ loss weight & 1.0 \\
    weight decay & 0.0 \\
    learning rate & 1e-4 \\
    lr schedule & constant \\
    batch size & 128 \\
    training epochs (continuous warm-up) & 10 \\
    training epochs (discrete learning) & 40 \\
    training epochs (alternate fine-tuning) & 10 \\
    \hline
    \end{tabular}
    \endgroup
    }
    \caption{Training hyper-parameters for our tokenizer.}
    \label{tab:training_tokenizer}
\end{minipage}
\hfill
\begin{minipage}{0.42\textwidth} 
    \centering
    \scalebox{1.0}{
    \begingroup
    \fontsize{12}{14}\selectfont
    \begin{tabular}{l|c}
    \hline
    Hyper-parameters & Configuration \\
    \hline
    optimizer & Adamw \\
    $\beta_1$ & 0.9 \\
    $\beta_2$ & 0.96 \\
    condition dropout & 0.1 \\
    attention dropout & 0.1 \\
    cross-entropy loss weight & 1.0 \\
    diffusion loss weight & 1.0 \\
    weight decay & 0.03 \\
    learning rate & 1e-4 \\
    lr schedule & cosine \\
    batch size (256$\times$256) & 1024 \\
    batch size (512$\times$512) & 1024 \\
    training steps (256$\times$256) & 200K \\
    training steps (512$\times$512) & 50K \\
    \hline
    \end{tabular}
    \endgroup
    }
    \caption{Training hyper-parameters for our generator.}
    \label{tab:training_generator}
\end{minipage}
\end{center}
\end{table}

\subsection{Additional Text-to-image Examples}
\label{subsec:text-to-image}

\begin{figure*}[t]
    \centering
\includegraphics[width=0.9\textwidth]{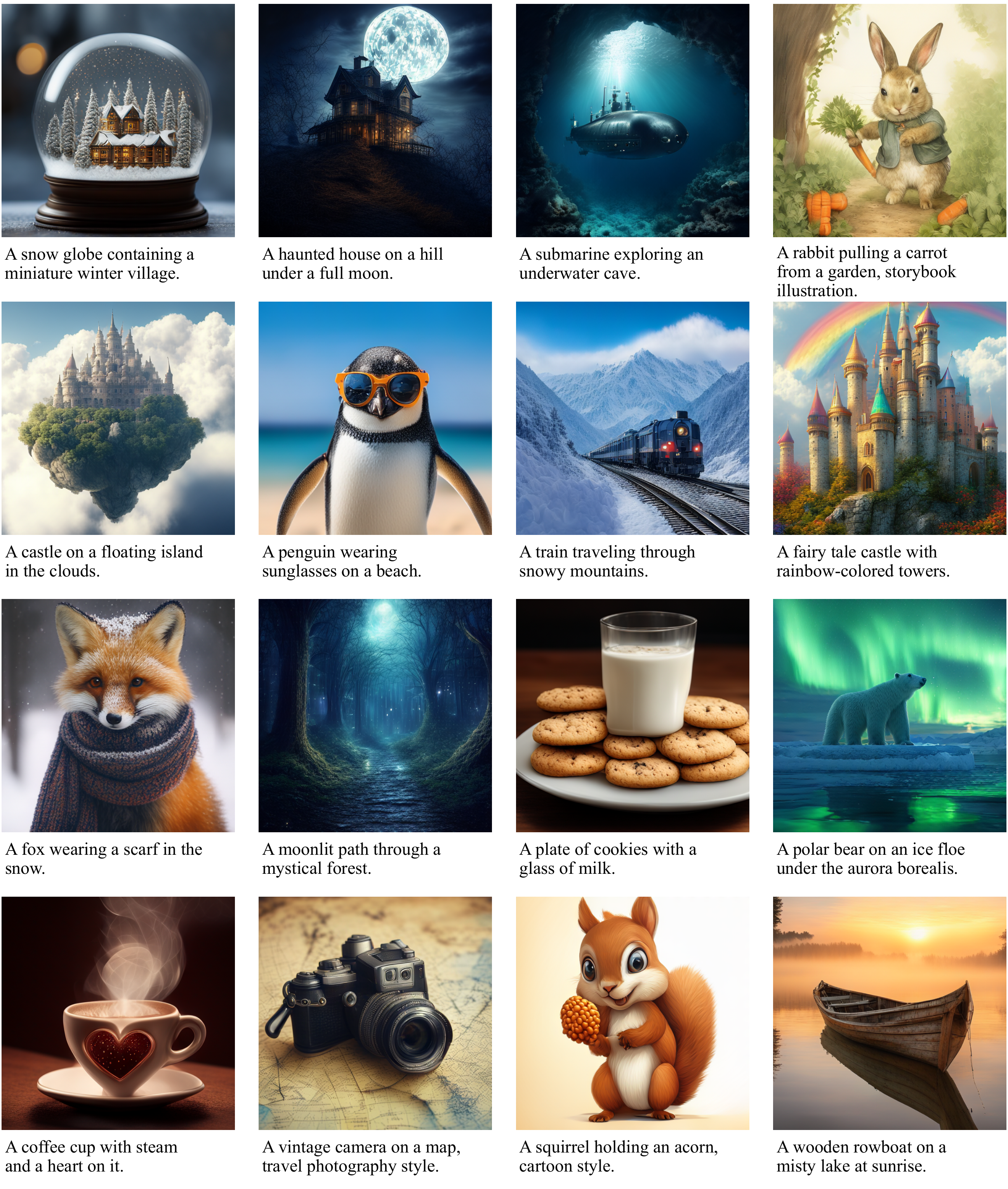}
    \caption{\textbf{Additional text-to-image generation results of \modelterm.}}
    \label{fig:additional_generation1}
\end{figure*}

\begin{figure*}[t]
    \centering
\includegraphics[width=0.9\textwidth]{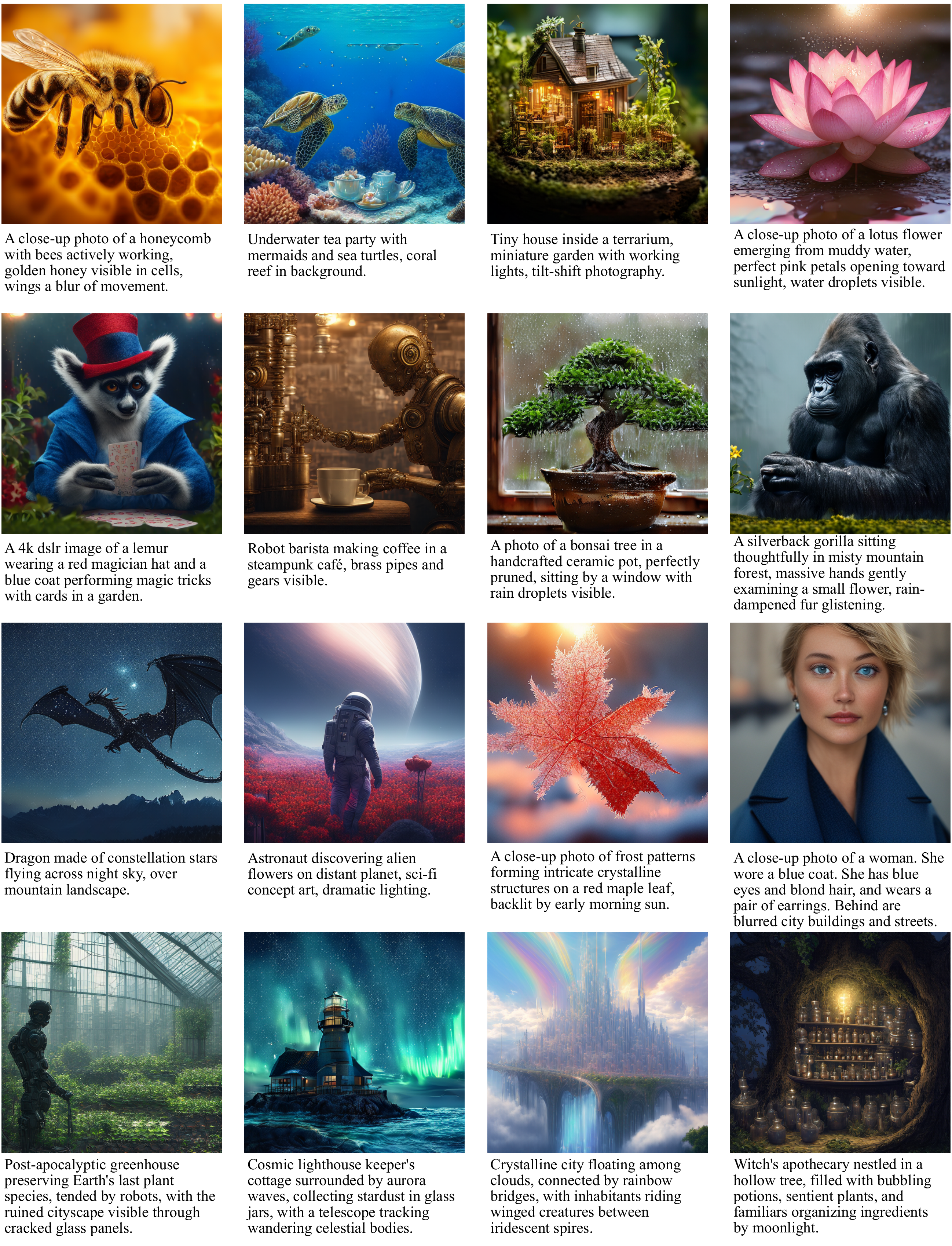}
    \caption{\textbf{Additional text-to-image generation results of \modelterm.}}
    \label{fig:additional_generation2}
\end{figure*}

\cref{fig:additional_generation1} and \cref{fig:additional_generation2} includes more qualitative examples of text-to-image generation results of \modelterm.

\subsection{Qualitative Comparison of \modelterm and discrete-only baseline.}
\label{subsec:qualitative_comparison}

We present qualitative comparison examples of images generated by \modelterm and the discrete-only baseline. From these examples, it is evident that the diffusion head and residual tokens significantly enhance image refinement, particularly in capturing fine details such as eyes and textures.

\begin{figure*}[t]
    \centering
\includegraphics[width=0.9\textwidth]{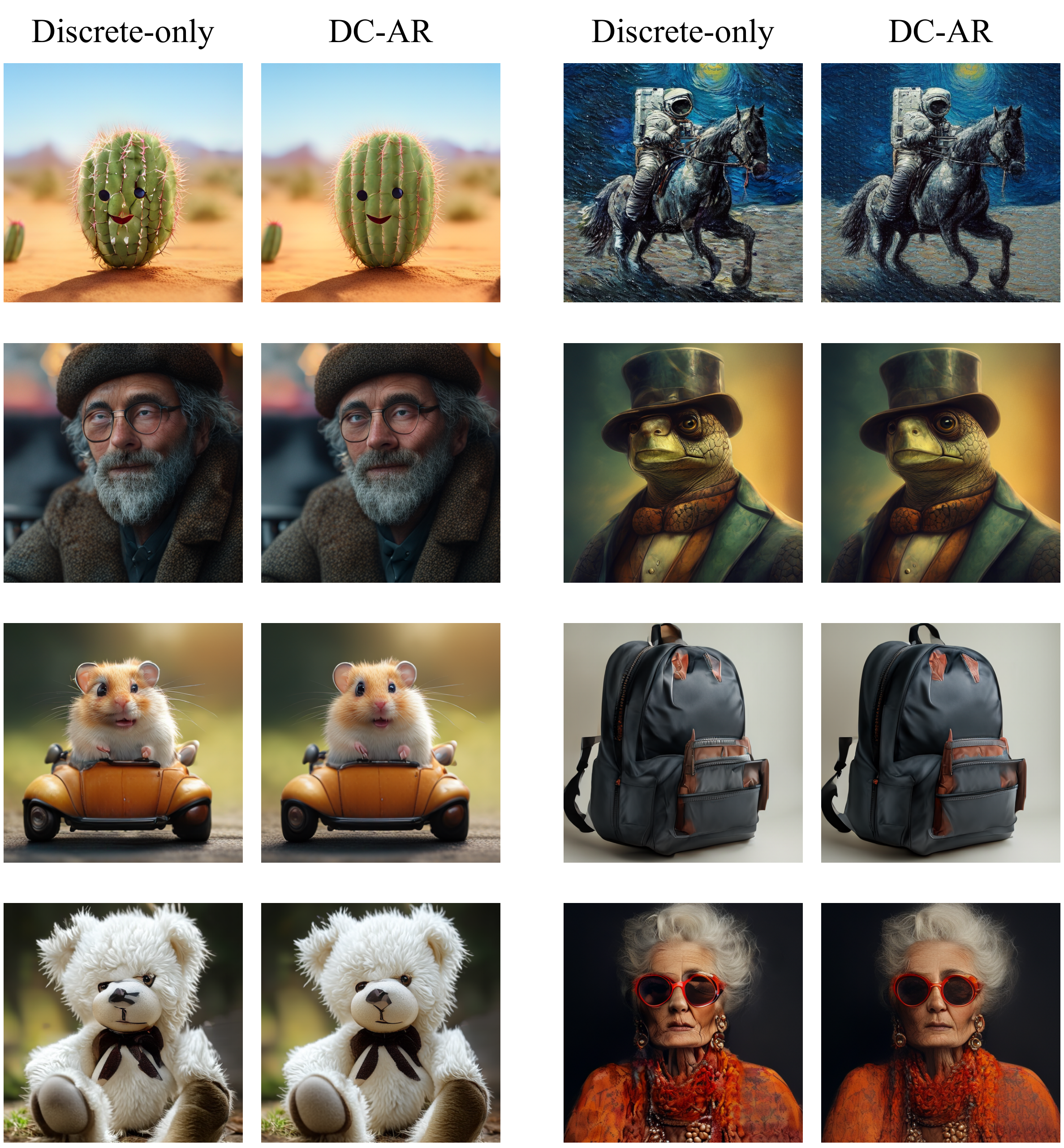}
    \caption{\textbf{Qualitative Comparison: Images Generated by \modelterm vs. the Discrete-Only Baseline.} For each pair of images, the left image is produced by the discrete-only baseline, while the right image is generated by \modelterm.}
    \label{fig:comparsion_discrete}
\end{figure*}

\subsection{Additional Experimental Results.}
\label{subsec:additional_experiments}

In this section, we provide some other experiments related to \modelterm.

\begin{figure}[t]
    \centering
    \begin{minipage}{0.45\textwidth} 
        \centering
        \includegraphics[width=\textwidth]{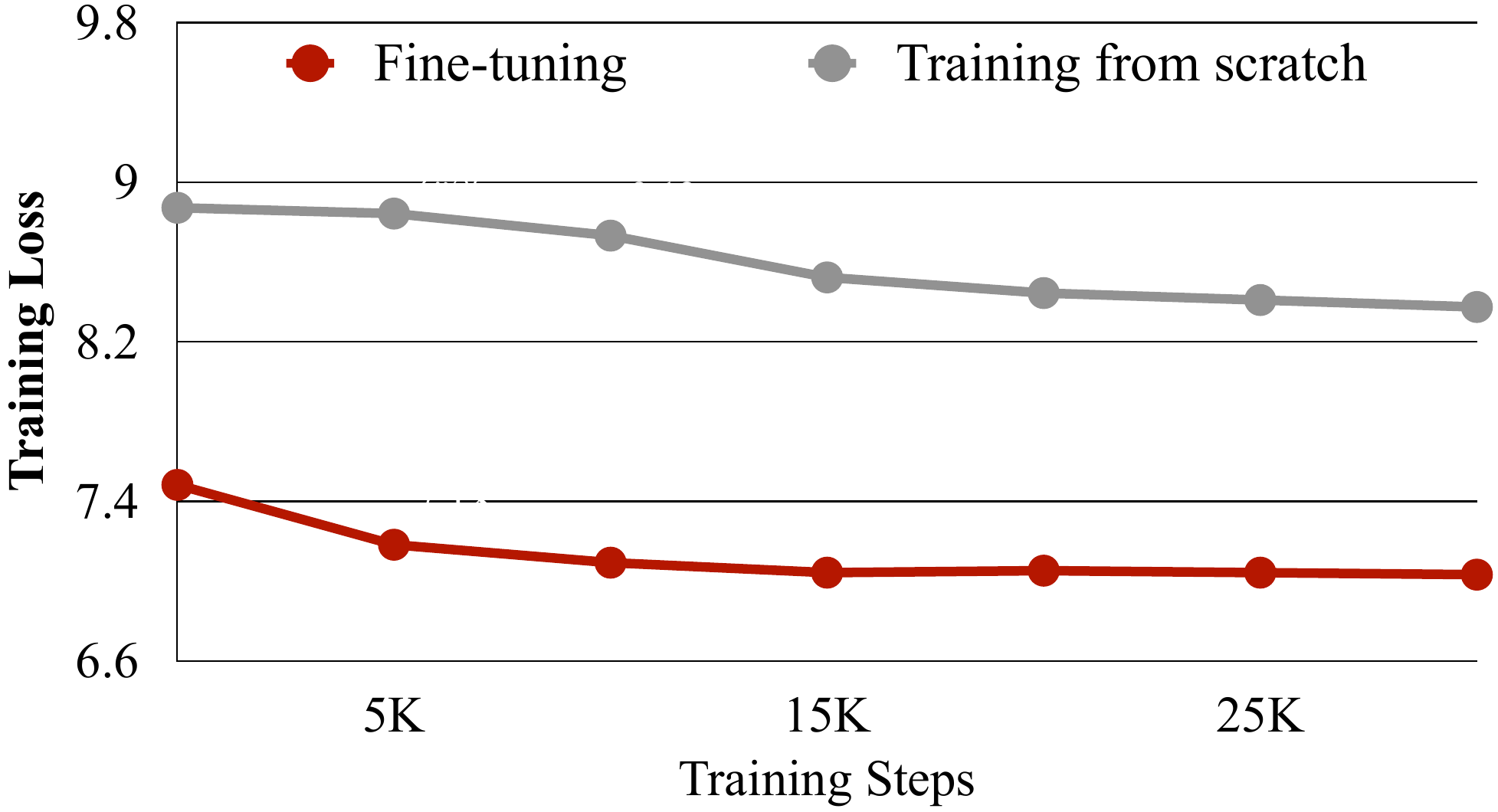}
        \caption{The resolution generalizability of \tokenizershort allows us to train a 512$\times$512 model by fine-tuning from a pre-trained 256$\times$256 model, achieving significantly faster convergence compared to training from scratch.}
        \label{fig:appendix_loss}
    \end{minipage}
    \hfill 
    \begin{minipage}{0.45\textwidth} 
        \centering
        \includegraphics[width=\textwidth]{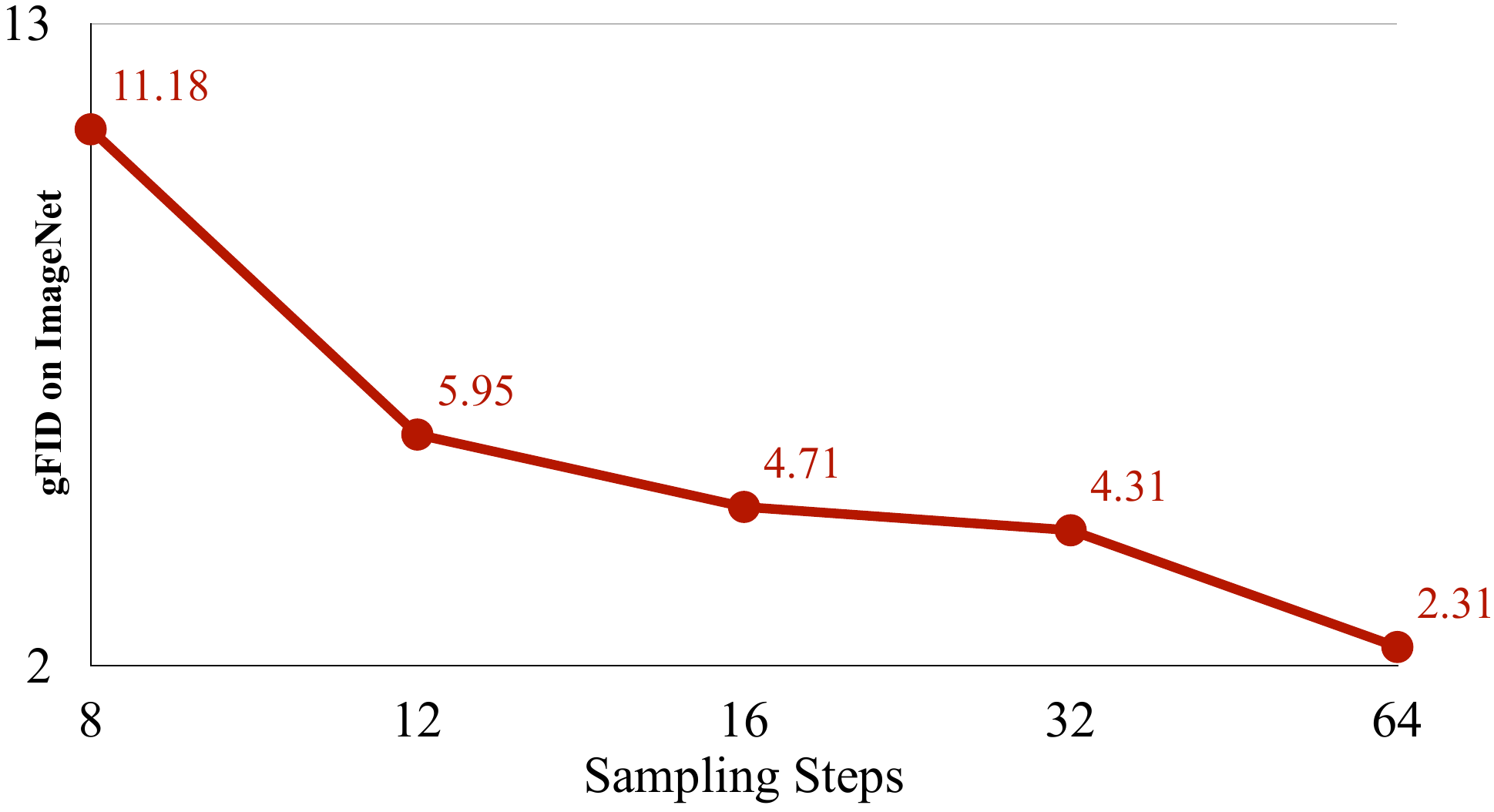}
        \caption{\textbf{gFID Results on ImageNet 256$\times$256 for MAR-B at Different Sampling Steps.} MAR-B requires 64 sampling steps to achieve its best performance, significantly lagging behind our method, which attains optimal performance in just 12 steps.}
        \label{fig:appendix_mar}
    \end{minipage}
\end{figure}

\myparagraph{Training Loss Curve: Fine-Tuning vs. Training from Scratch.} \cref{fig:appendix_loss} illustrates the training loss curves for fine-tuning and training from scratch on 512$\times$512 models over the first 30K steps. It is evident that fine-tuning from a pre-trained 256$\times$256 model enables the 512$\times$512 model to converge significantly faster than training from scratch.

\myparagraph{Sampling Step Requirements for MAR.} Our primary motivation for adopting a hybrid generation framework, rather than following MAR's paradigm of exclusively using continuous tokens, stems from the observation that MAR typically requires a large number of steps to achieve optimal performance. This is demonstrated in \cref{fig:ablation_mar}, where we evaluate the official MAR-B model for class-conditional image generation on ImageNet at 256$\times$256 resolution. Despite the image token sequence length being 256, MAR-B requires 64 steps to reach its optimal performance, resulting in a substantial inference cost. In contrast, \modelterm achieves optimal performance in just 12 steps, making it significantly more efficient during sampling.

\subsection{Discussion of \modelterm and Related Works.}

As a novel autoregressive image generation framework, \modelterm draws inspiration from several related works in the field while introducing significant innovations that distinguish it from each of them.

\myparagraph{Difference with MaskGen \cite{kim2025democratizing}.} Both MaskGen and \modelterm adopt the masked autoregressive generation paradigm for text-to-image generation and employ an image tokenizer with a high compression ratio for efficient generation. However, their technical approaches to building the tokenizer and generator differ substantially. On the tokenizer side, MaskGen follows the recent trend of using a 1D compact tokenizer to achieve a high compression ratio. However, a major limitation of such 1D tokenizers is their lack of generalizability across different resolutions. Consequently, MaskGen must train separate tokenizers and generators from scratch for each resolution, leading to significantly higher training costs, especially for resolutions of 512$\times$512 or higher. In contrast, \modelterm utilizes a single tokenizer trained on 256$\times$256 images for all resolutions and fine-tunes the generator for higher resolutions from a pre-trained low-resolution model, resulting in much greater efficiency. On the generator side, MaskGen combines the MaskGIT paradigm for discrete token generation with the MAR paradigm for continuous token generation. In contrast, \modelterm introduces a novel hybrid generation framework that leverages the superior representation capability of continuous tokens while maintaining the high inference speed of discrete tokens.

\myparagraph{Difference with HART \cite{tang2024hart}.} HART proposes the idea of hybrid tokenization, using a transformer model to generate discrete tokens and a lightweight diffusion head to generate continuous tokens. While \modelterm inherits these ideas, it adapts them in a fundamentally different setting. HART follows the VAR paradigm, which generates images through progressive next-scale refinement. In contrast, \modelterm adopts the MaskGIT paradigm, which generates images through progressive unmasking. Although the VAR paradigm is widely recognized for its high generation quality and speed, we believe the MaskGIT paradigm offers unique advantages, including fewer tokens (VAR requires additional tokens due to its multi-scale tokenization design) and a natural suitability for image editing tasks. Building on this foundation, \modelterm introduces novel methods, such as a single-scale hybrid tokenizer with a 32$\times$ compression ratio (via our three-stage adaptation strategy) and an efficient hybrid generation framework that extends MaskGIT (via our discrete token-dominated generation pipeline). Notably, in the results section, we do not include comparisons with VAR-based methods, as we aim to focus the discussion on how \modelterm advances the MaskGIT paradigm. In future work, we plan to explore adapting our approach to the VAR paradigm to design even more effective generation frameworks.

\end{document}